\documentclass{article}

\usepackage{iclr2023_conference}

\usepackage[utf8]{inputenc} % allow utf-8 input
\usepackage[T1]{fontenc}    % use 8-bit T1 fonts
\usepackage{hyperref}       % hyperlinks
\usepackage{url}            % simple URL typesetting
\usepackage{booktabs}       % professional-quality tables
\usepackage{amsfonts}       % blackboard math symbols
\usepackage{nicefrac}       % compact symbols for 1/2, etc.
\usepackage{microtype}      % microtypography
\usepackage{xcolor}         % colors
\usepackage{graphicx}       % use figures
\usepackage{caption}
\usepackage{subcaption}
\usepackage{amsmath}
\usepackage{adjustbox}
\usepackage{hyphenat}

\newcommand{\comment}[1]{}  %Comments 

\iclrfinaltrue

\title{\textbf{\nohyphens{MAHTM: A Multi-Agent Framework for Hierarchical Transactive Microgrids}}}

\author{Nicolas M. Cuadrado 
\thanks{These authors contributed equally to this work.} 
\\
MBZUAI, UAE \\
\texttt{nicolas.avila@mbzuai.ac.ae} \\ 
\And
Roberto A. Gutiérrez $^*$
\\
MBZUAI, UAE\\
\texttt{roberto.guillen@mbzuai.ac.ae}  \\ 
\And
Yongli Zhu \\
Texas A\&M University, USA \\
\texttt{yzhu16@vols.utk.edu} $\qquad\qquad\quad\qquad $  \\
\And
Martin Taká\v{c}
\\
MBZUAI, UAE\\
\texttt{Takac.MT@gmail.com}
}

\usepackage{array}
\usepackage{arydshln}
\begin{document}

\maketitle

\begin{abstract}
    Integrating variable renewable energy into the grid has posed challenges to system operators in achieving optimal trade-offs among energy availability, cost affordability, and pollution controllability. This paper proposes a multi-agent reinforcement learning framework for managing energy transactions in microgrids. The framework addresses the challenges above: it seeks to optimize the usage of available resources by minimizing the carbon footprint while benefiting all stakeholders. The proposed architecture consists of three layers of agents, each pursuing different objectives. The first layer, comprised of prosumers and consumers, minimizes the total energy cost. The other two layers control the energy price to decrease the carbon impact while balancing the consumption and production of both renewable and conventional energy. This framework also takes into account fluctuations in energy demand and supply. 
    % Renewables stochastic problem
    % Multi-objective, greedy, and collective goals
    % Using RL, VRE (Variable Renewable Energy) 
    % Hierarchical approach
    % Climate change impact

\end{abstract}

\section{Introduction}

As technology and urban areas continue to grow, the demand for energy increases and is expected to continue to be high. Because of this, the world is moving towards greener options, increasing the demand for renewable energy from industry and residential consumers \cite{LEIBOWICZ20181311}. Between renewable energies, we have solar, wind, tidal, hydropower, and bio-energy. The challenge behind lays not only in the generation but guaranteeing there is enough to supply the demand since the generation of renewable energy is inherently stochastic (depends on multiple climate factors).

There is a need to adapt to the randomness of the situation, which can be solved by creating specific energy systems controlled by machine learning models, which optimize the usage of the available resources \cite{VAZQUEZCANTELI2019243}. For example, the concept of ``smart transactive grids'' has been proposed to organize the demand and production of energy in communities. The idea is to create an intelligent system that uses different energy sources to supply the demand with minimal human intervention. At the same time, it provides the opportunity to sell any surplus energy produced.

Some previous work leveraged Reinforcement Learning (RL) to create technologies that enable transactive microgrids. In \cite{multiagent_buildings}, an approach using different RL agents is proposed, where the distributions of agents are different: one agent is used for particular computation (e.g., optimization), called Service Agent; the other two agents collect meteorological information, and forecast the power output based on the specified type of energy (solar, wind, etc.). In the design of its energy management system, one battery is shared across the residential households, and all the agents communicate with a ``central coordinator agent''. In \cite{MARLISA}, the authors proposed a Multi-Agent Reinforcement Learning (MARL) approach consisting of agents sharing two variables and following a leader-follower schema to manage energy demand and generation. They also proposed a specific reward function with greedy and collective goals to incentivize the agents to work together as a community.

Other MARL approaches could be relevant to solve similar issues. In COLA \cite{COLA}, a consensus learning approach is proposed, which is inspired by the DINO (Distillation with No Labels) \cite{caron2021emerging}. In DINO's method, a student network is created to predict the teacher network results to simplify the self-supervised training process for methods that (originally) require centralized training and decentralized inference. Authors of \cite{COMA} proposed the Counterfactual Multi-Agent (COMA) policy gradients. In this work, they propose an architecture with a centralized critic to estimate the action-value function and decentralized actors to learn the optimal policy for each agent. The main innovation in this approach is introducing a counterfactual baseline that allows each agent to compare its current action contribution to the global reward with all the other possible actions. This is a way to deal with the problem of \textit{credit assignment} in the multi-agent context, which happens when the agents do not consider their contribution to a collaborative objective.

This paper proposes and develops a Multi-Agent Hierarchical Framework for Transactive Microgrids (MAHTM). The framework considers \textbf{the minimization of the carbon footprint} during the multi-agent learning process to tackle the challenges of climate change.

\section{Methods}

\begin{figure}[ht]
        \includegraphics[width=\textwidth]{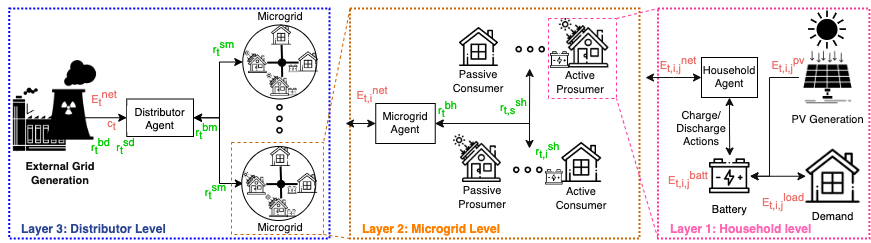}
    \caption{Illustration on the Three Layer Architecture. The solid lines represent the energy and information flows.}
    \label{fig:architecture} 
\end{figure}

We propose a three-layer hierarchical RL architecture, as shown in Figure. \ref{fig:architecture}. Each layer owns a set of agents with different objectives, pursued greedily. In our framework, we denote a set of $G$ microgrids as $M = \{ m_1, m_2, \dots, m_i, \dots, m_G \}$; we denote a group of $D_i$ households belonging to a microgrid $i$ as $H_i = \{ h_{i,1}, h_{i,2}, \dots, h_{i,j}, \dots, h_{i,D_i}\}$, the current time step is denoted as $t$.

\begin{table} 
    \renewcommand*{\arraystretch}{1.}
    \centering
    \begin{tabular}{@{}lccccc@{}}
        \toprule
         &
          \textbf{L1: Household} &
          \textbf{L2: Microgrid} &
          \textbf{L3: Distributor} &
          \textbf{Type} &
          \textbf{Unit} \\ 
          \midrule
        Net & $E^{\text{net}}_{t,i,j}$ & $E^{\text{net}}_{t,i}$ & $E^{\text{net}}_{t}$ & Energy      & Wh     \\ \hdashline
        Demand & $E^{\text{load}}_{t,i,j}$ & - & - & Energy      & Wh     \\ \hdashline
        PV Gen & $E^{\text{pv}}_{t,i,j}$ & - & - & Energy      & Wh     \\ \hdashline
        Battery & $E^{\text{batt}}_{t,i,j}$ & - & - & Energy      & Wh     \\ \hdashline
        Shortage & $E^{\text{st}}_{t,i,j}$ & $E^{\text{st}}_{t,i}$ & $E^{\text{st}}_{t}$ & Energy      & Wh     \\ \hdashline
        Surplus & $E^{\text{sp}}_{t,i,j}$ & $E^{\text{sp}}_{t,i}$ & $E^{\text{sp}}_{t}$ & Energy      & Wh     \\ \hdashline
        L1 Import & $E^{imp_1}_{t,i,j}$ & - & - & Energy      & Wh     \\ \hdashline
        L1 Export & $E^{exp_1}_{t,i,j}$ & - & - & Energy      & Wh     \\ \hdashline
        L2 Import & $E^{imp_2}_{t,i,j}$ & $E^{imp_2}_{t,i}$ & -  & Energy      & Wh     \\ \hdashline
        L2 Export & $E^{exp_2}_{t,i,j}$ & $E^{imp_2}_{t,i}$ & - & Energy      & Wh     \\ \hdashline
        L3 Import & $E^{imp_3}_{t,i,j}$ & $E^{imp_3}_{t,i}$ & $E^{imp_3}_{t}$ & Energy      & Wh     \\ \hdashline
        L3 Export & $E^{exp_3}_{t,i,j}$ & $E^{imp_3}_{t,i}$ & $E^{imp_3}_{t}$ & Energy      & Wh     \\ \hdashline
        Emission & - & - & $c_t$ & GHG & CO$_2$/Wh \\ \hdashline
        Sell & $r^{\text{sh}}_{t,i}$ & $r^{\text{sm}}_t$ & $r^{\text{sd}}_t$ & Price       & \$/Wh  \\  \hdashline
        Buy & $r^{\text{bh}}_{t,i}$ & $r^{\text{bm}}_t$ & $r^{\text{bd}}_t$ & Price       & \$/Wh  \\ 
        \bottomrule
    \end{tabular}
    \vskip-5pt
    \label{tab:symbols}
    \caption{Table of defined symbols.}
\end{table}

\subsection{First layer: household}

In this layer, there are four different cases: 1) households that have no access to any energy asset being only able to consume (''passive consumers''); 2)  households that have access to Photovoltaic (PV) panels to produce electricity during day-hours (''passive prosumers''); 3) prosumer households that have access to batteries which allow them to have energy dispatch capabilities and PV generation (''active prosumers''); and  4) consumer households who also have access to energy storage which provide them the potential to sell surplus energy back to the microgrid (''active consumers''). Households without batteries (''passive consumers'' or ''passive prosumers'') do not need to execute control actions as they do not have such capabilities to react to energy fluctuations (e.g., due to weather variation). In contrast, those ''actionable'' agents will determine how to charge and discharge the batteries and how to alter the demand and supply in the microgrid. Based on the above logic, the equations of this layer are as follows: 
\begin{align}
        E^{\text{st}}_{t,i,j} &= E^{imp_1}_{t,i,j} + E^{imp_2}_{t,i,j} + E^{imp_3}_{t,i,j}
    \label{eq:shortage}
    \\
        E^{\text{sp}}_{t,i,j} &= E^{exp_1}_{t,i,j} + E^{exp_2}_{t,i,j} + E^{exp_3}_{t,i,j}
    \label{eq:surplus}
    \\
        E^{\text{net}}_{t,i,j} &= E^{\text{st}}_{t,i,j} - E^{\text{sp}}_{t,i,j} = E^{\text{load}}_{t,i,j} - E^{\text{pv}}_{t,i,j} \pm E^{\text{batt}}_{t,i,j}
    \label{eq:net_equilibrium}
\end{align}
In the case of consumer households with no PV panel, the generation $E^{\text{pv}}_{t,i,j} = 0$. When $E^{\text{net}}_{t,i,j} \geq 0$ (called ``shortage'' state), it means there is extra energy needed from external sources (e.g., retailers or other households). When $E^{\text{net}}_{t,i,j} < 0$ (called ``surplus'' state), there is surplus energy available to sell back to the external power grid or other households in shortage. The Equation \eqref{eq:net_equilibrium} presents a constraint that should be satisfied as it is impossible to have both scenarios simultaneously. Finally, we define the objective function of this layer:
\begin{equation}
    \min
    \left\{
        \begin{array}{lr} 
            E^{imp_3}_{t,i,j} (r^{\text{sd}}_t + c_t) + E^{imp_2}_{t,i,j} r^{\text{sm}}_{t,i} + E^{imp_1}_{t,i,j} r^{\text{sh}}_{t,i}, & \text{if} ~~~ E^{\text{net}}_{t,i,j} \geq 0 , \\
            E^{exp_3}_{t,i,j} r^{\text{bd}}_t + E^{exp_2}_{t,i,j} r^{\text{bm}}_t + E^{exp_1}_{t,i,j} r^{\text{bh}}_{t,i}, & \text{if} ~~~ E^{\text{net}}_{t,i,j} < 0 .
        \end{array}
    \right.
    \label{eq:obj_layer1}
\end{equation}

\subsection{Second layer: microgrid}

In this layer, an agent defines the prices $r^{\text{sh}}_{t,i}$ and $r^{\text{bh}}_{t,i}$. Its objective is to maximize the use of local energy in a microgrid by defining the pricing policy for local transactions. We described it using the following equations:
\begin{align}
        E^{\text{st}}_{t,i} &= E^{imp_2}_{t,i} + E^{imp_3}_{t,i},&
        ~ E^{imp_2}_{t,i} &= \textstyle{\sum}_{j} E^{imp_2}_{t,i,j} ,
        & E^{imp_3}_{t,i} &= \textstyle{\sum}_{j} E^{imp_3}_{t,i,j}, 
     \label{eq:total_imported}
     \\
        E^{\text{sp}}_{t,i} &= E^{exp_2}_{t,i} + E^{exp_3}_{t,i},&  
        E^{exp_2}_{t,i} &= \textstyle{\sum}_{j} E^{exp_2}_{t,i,j},
        &
        E^{exp_3}_{t,i} &= \textstyle{\sum}_{j} E^{exp_3}_{t,i,j}, 
    \label{eq:total_exported}
    \\
        E^{\text{net}}_{t,i} &= E^{\text{st}}_{t,i} - E^{\text{sp}}_{t,i}.
    \label{eq:net_l2}
\end{align}
A microgrid will experience an (energy) shortage state when the local energy is insufficient to cover the internal demand and experience an (energy) surplus state when the distributed generation surpasses the internal demand. In the first case, a microgrid could access energy available in other microgrids. In the second case, it could sell energy to other microgrids experiencing a shortage. If energy is unavailable/over-produced at the current microgrid layer, it will be imported or exported to the third layer. With this, we can define this layer's objective function:
\begin{align}
    \min \left\{
    \begin{array}{lr} 
        E^{imp_3}_{t,i} (r^{\text{sd}}_t + c_t) + E^{imp_2}_{t,i} r^{\text{sm}}_t, & \text{if} ~~ E^{\text{net}}_{t,i} \geq 0  ,\\
        E^{exp_3}_{t,i} r^{\text{bd}}_t + E^{exp_2}_{t,i} r^{\text{bm}}_t, & \text{if} ~~ E^{\text{net}}_{t,i} < 0.
    \end{array}
    \right.
    \label{eq:obj_layer2}
\end{align}

\subsection{Third layer: distributor}

In this layer, the agent tries to shape the overall load among the multiple microgrids, enabling energy trading and simultaneously \textit{minimizing the carbon footprint} by setting the buy ($r^{\text{bm}}_t$) and sell ($r^{\text{sm}}_t$) prices among the microgrids. The prices for selling energy ($r^{\text{sd}}_t$) and accepting surplus ($r^{\text{bd}}_t$) from the microgrids are not controlled in this layer and are treated as external inputs (from the previous layer). To define the objective function of the distributor, we need first to define the following:
\begin{equation}
    \begin{aligned}
        E^{\text{st}}_{t} = E^{imp_3}_{t} = \textstyle{\sum}_{i} E^{imp_3}_{t,i}
    \end{aligned}
    \label{eq:total_imported_mg}
\end{equation}
\begin{equation}
    \begin{aligned}
        E^{\text{sp}}_{t} = E^{exp_3}_{t} = \textstyle{\sum}_{i} E^{exp_3}_{t,i}
    \end{aligned}
    \label{eq:total_exported_mg}
\end{equation}
\begin{equation}
    \begin{aligned}
        E^{\text{net}}_{t} = E^{\text{st}}_{t} - E^{\text{sp}}_{t} = E^{imp_3}_{t} - E^{exp_3}_{t}
    \end{aligned}
    \label{eq:net_l3}
\end{equation}
Then, we can define the distributor's objective function as follows:
\begin{equation}
    \min \left\{
    \begin{array}{lr} 
        E^{imp_3}_{t} (r^{\text{sd}}_t + c_t), & \text{if} ~~ E^{\text{net}}_{t} \geq 0 , \\
        E^{exp_3}_{t} r^{\text{bd}}_t, & \text{if} ~~ E^{\text{net}}_{t} < 0.
    \end{array}
    \right.
    \label{eq:obj_layer3}
\end{equation}
In addition, we assume there is only one distributor and that the energy consumed within or between microgrids has negligible carbon impact. We also implemented a simple local energy market based on the physical distance between the household and the microgrids.

\section{Results and Analysis}

\subsection{Experimental setup}

We configured our environment (in OpenAI Gym) to present different sets of households for training, validation, and testing. Detail about the precise attributes of the dataset is present in the appendix. A critical difference from existing work like \cite{COLA} is the possibility of enabling and disabling the stochasticity in our setup.

\subsection{Model performance}

We propose a performance metric based on how energy cost and carbon impact are improved by optimally managing distributed storage. The metric measures the scenario without batteries against the use of our hierarchical control. Each household contributes to the metric individually, and the upper levels aggregate them to have microgrid-level and distributor-level performance.
%TODO Review update
\begin{table}[htbp]
    \centering
    \begin{tabular}{l r r r }
    \toprule
  & CVXPY & MAHTM & COMA \\ \midrule
        Train reward & \textbf{-0.915} & -0.993 & -1.3 \\  \hdashline
        Train price score & \textbf{-0.103} & -0.097 & 0.35 \\  \hdashline
        Train emission score & \textbf{-0.223} & -0.1522 & 0.35 \\  \hdashline
        Train time & \textbf{0.9s} & 10m & 2h \\  \hdashline
        Test price score & \textbf{-0.0889} & -0.064 & 0.0625 \\  \hdashline
        Test emission score & \textbf{-0.19} & -0.097 & 0.0625 \\ \bottomrule
    \end{tabular}
    \vskip-5pt
    \caption{Average performance of households (lower is better, except for reward).}
    \label{tab:results}
\end{table}

Table \ref{tab:results} presents our current empirical results comparing the optimal solution for our scenario using a linear solver (CVXPY)\cite{cvxpy}, our framework, and COMA (one of the state-of-the-art MARL algorithms). One of the things to highlight about our approach is its training speed and simplicity versus COMA, which is very sensitive to hyperparameter tuning. Our framework reached solutions very close to the optimal within a reasonable training time.
%TODO Review update

\section{Conclusion}

The proposed framework systematically applies the MARL technique to transactive microgrids. The results are compared with one classic MARL algorithm. A customized OpenAI Gym environment was also created to serve as the test bench for this work. Our framework can help the development of local renewable energy markets, fostering emission reduction and more consumer engagement.
\\
\\
(The source code and demo files have been anonymized and are available in \href{https://github.com/nicosquare/rl-energy-management}{\textbf{this repository}} link.)

\bibliographystyle{plainnat}
\bibliography{references.bib}

\newpage

\section*{Appendix}

\subsection*{Algorithms}

\subsubsection*{First layer: Policy Gradient (PG), Advantage Actor-Critic (A2C)}

In this approach, the objective of our agent is to maximize the probability of having the trajectories that show the higher sum reward. It is defined as:

\begin{equation}
    J(\theta) = E_{\pi_\theta}\left[\sum_t \gamma^t r_t\right]
    \label{eq:rl_reward}
\end{equation}

It can be understood as the expected sum of the discounted rewards obtained by completing one episode following a defined policy $\pi_\theta$. The factor $\gamma$ helps prevent the sum from going infinite and gives more relevance to the rewards obtained in the short term. The whole idea of this RL method is to maximize \ref{eq:rl_reward} using stochastic gradient ascent. By using the definition of expectation, we can define the policy gradient as:

\begin{equation}
    \nabla_\theta J(\theta) = E_{\tau \sim \pi_\theta(\tau)}\left[\left(\sum_{t=1}^T \nabla_\theta log \pi_\theta(a_t|s_t)\right)\left(\sum_{t=1}^T r(s_t,a_t)\right)\right]
    \label{eq:rl_policy_gradient}
\end{equation}

In A2C, we use an estimator (a neural network) to represent the policy $\pi_\theta(a_t|s_t)$, named Actor. The actor will map the states to the actions and learn the optimal ones. Its training follows the next steps:

\begin{itemize}
    \item Sample $i$ trajectories $\tau^i$ using the actor policy.
    \item Assuming the policy gradient definition in \ref{eq:rl_policy_gradient}.
    \item Updating the weights $\theta$ of the policy as follows: $\theta \leftarrow \theta + \alpha \nabla_\theta J(\theta)$.
\end{itemize}

Sequentially running multiple trajectories is a long process. For that reason, batch training is generally implemented to speed up the learning of the policy estimator. By doing so, the exploration speed increases, modifying the equation \ref{eq:rl_policy_gradient} as follows:

\begin{equation}
    \nabla_\theta J(\theta) \approx \frac{1}{N} \sum_{i=1}^N \sum_{t=1}^T \left[\nabla_\theta log \pi_\theta(a_t^i|s_t^i)\left(\sum_{t=1}^T r(s_t^i,a_t^i)\right)\right]
    \label{eq:rl_batch_policy_gradient}
\end{equation}

However, by doing so, we add an issue: The variance of $\nabla_\theta J(\theta)$ increases. To help solve this, the advantage function was introduced. First, we start by understanding that the term $\left(\sum_{t=1}^T r(s_t^i,a_t^i)\right)$ is the $Q(s,a)$ function, as it represents the expected reward we can get from doing an action $a_t$ while in state $s_t$. Finding a value $V$ independent of the neural network parameters $\theta$, we can subtract it from the $Q$ function to re-calibrate the rewards towards the average action. Thus, the advantage function is defined as:

\begin{equation}
    A^\pi(s_t,a_t) = Q^\pi(s_t,a_t) - V^\pi(s_t)
    \label{eq:rl_advantage_a2c}
\end{equation}

The algorithm A2C gets its name from the use of the advantage function\ref{eq:rl_advantage_a2c}, and the addition of an extra neural network (the Critic) that approximates $V^\pi(s_t)$ and will be trained with the experienced $Q^\pi(s_t,a_t)$. In other words, the critic evaluates the actions taken by the actor and approximates the corresponding values. 
    % 2nd layer, local microgrid multi-agent, manage prices, still under discussion, COLA, 

\subsubsection*{Multi-agent RL (MARL)}

This is the simplest way of implementing policy gradients in a multi-agent configuration. In this case, each agent has its actor and critic, interacting with an agent-specific action and observation history. This was first introduced in \cite{Tan1993MultiAgentRL} with a Q-learning algorithm. When using the same principle with an AC algorithm, it is called an Independent Actor-Critic (IAC) as explained in \cite{COMA}.

In this approach, all the agents' neural networks share parameters. Thus, only one agent and one critic are learned in the end. However, each agent has access to different observations and attributes associated with the household, allowing them to take different actions. This method helps RL agents with similar tasks to learn faster and better. %\cite{Tan1993MultiAgentRL} 

%MARL
Expanding on the equation (\ref{eq:rl_reward}) for the single-agent case, the equation (\ref{eq:marl})  is the equivalent for the multi-agent case. Where the generalization of the Markov decision process is the stochastic game, the state transitions and the rewards of the agents $r_{i,t+1}$ result from their joint actions.
\begin{equation}
    J(\theta)=\mathrm{E}\left[\sum_{t}^{} \gamma^t r_{i, t} \mid x_0=x, h\right]
    \label{eq:marl}
\end{equation}

% OG from paper
% \begin{equation}
%     R_i^h(x)=\mathrm{E}\left\{\sum_{k=0}^{\infty} \gamma^k r_{i, k+1} \mid x_0=x, h\right\}
%     \label{eq:marl}
% \end{equation}

\subsection*{Environment}
% Add open ai gym reference 
Using the OpenAI Gym toolkit \cite{openaigym} as a wrapper for an environment that uses data generated synthetically. It standardizes the evaluation of diverse RL agents. Our environment includes stochastic energy generation, agent energy market participation, realistic battery representation, and a diversified set of household demand profiles.

\subsubsection*{Dataset}
% Review the first sentence

\begin{figure}[ht]
    \centering
    \begin{subfigure}{0.45\textwidth}
        \includegraphics[width=\textwidth]{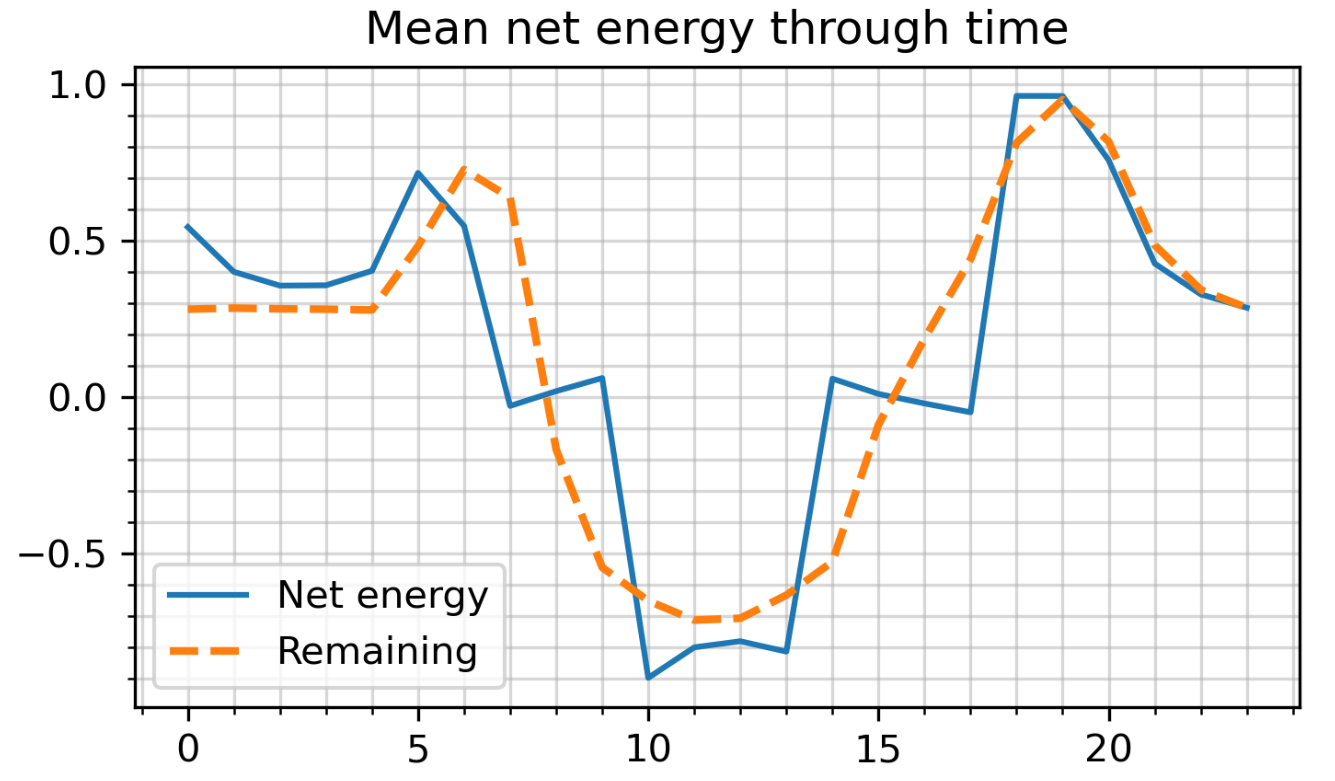}
        \caption{Family.}
        \label{fig:cvx_family} 
        \includegraphics[width=\textwidth]{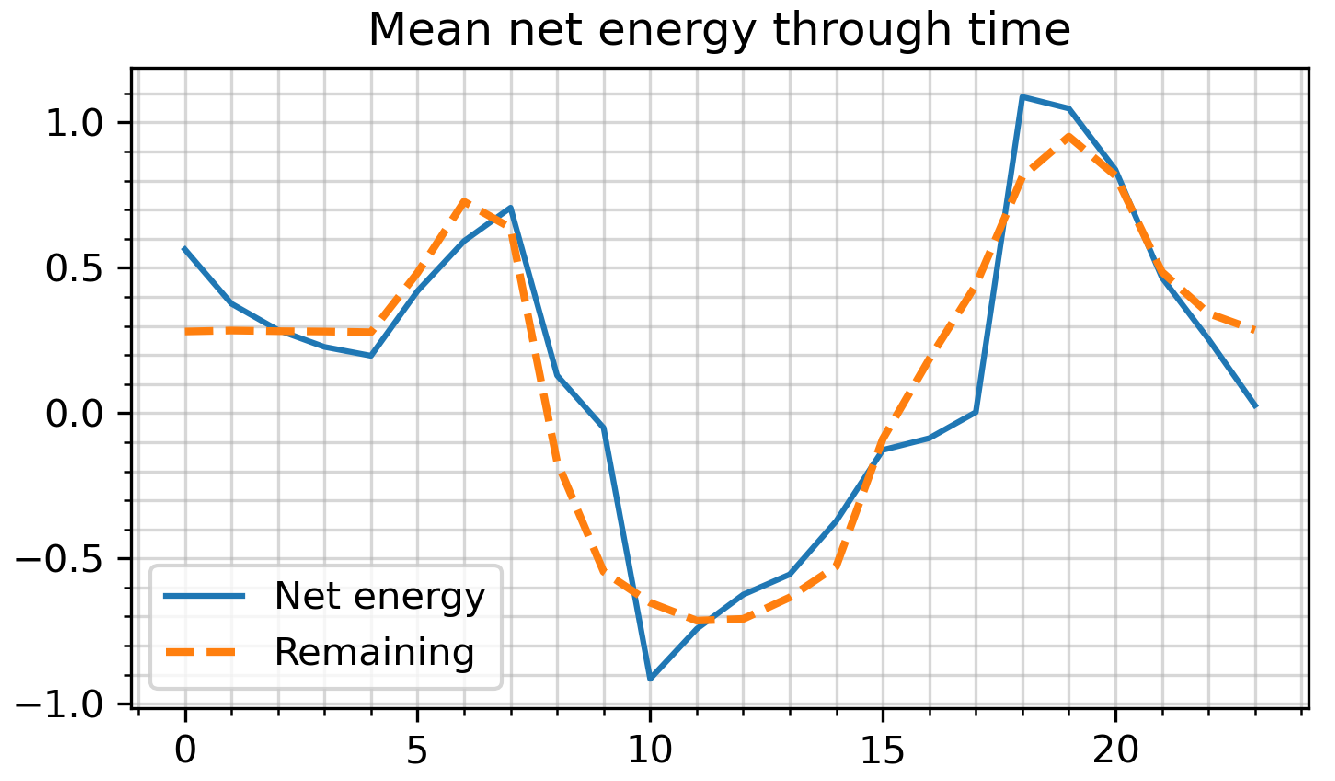}
        \caption{Teenagers.}
        \label{fig:cvx_teenagers} 
    \end{subfigure}
    \begin{subfigure}{0.45\textwidth}
        \includegraphics[width=\textwidth]{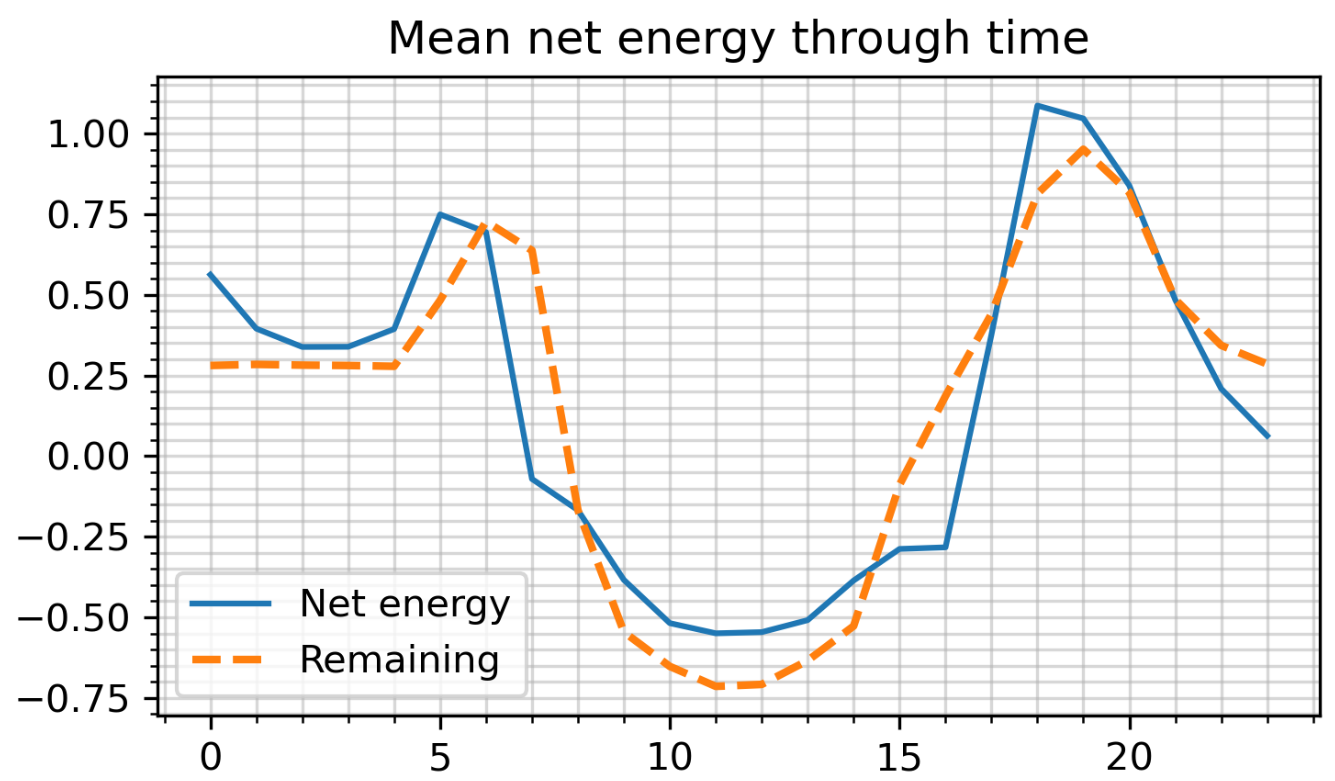}
        \caption{Business.}
        \label{fig:cvx_business} 
        \includegraphics[width=\textwidth]{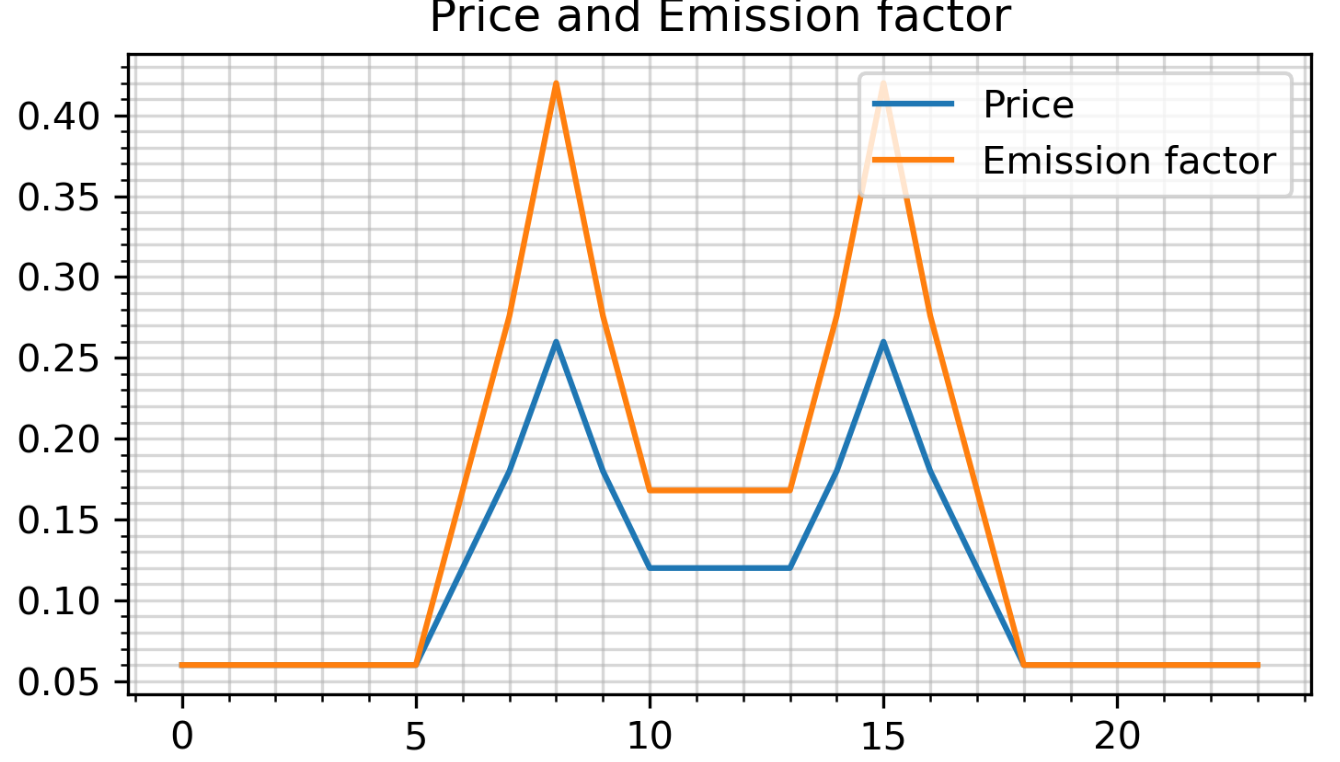}
        \caption{Price and emissions.}
        \label{fig:env_price_emissions} 
    \end{subfigure}
    \caption{Sample of solutions with solver (CVXPY).}
    \label{fig:cvx_optimize}
\label{fig:solution_opt}
\end{figure}

The data used in the current environment was synthetically generated based on real-life data. For this work, we defined 24 steps representing the hours within a day, with the possibility of extending to more steps if required. There are three different demand profiles, each representing a specific use: \textbf{family}, with demand peaks in the morning and early afternoon; \textbf{teenagers}, with peaks late in the afternoon till early morning; \textbf{house business}, with high energy usage in the middle of the day (refer to Figure \ref{fig:solution_opt} to see the described trends). These non-shiftable demands are generated with noise, different energy baselines, and a dependency on stochastic variables such as temperature. Grid energy cost and carbon footprint are defined by two different sources, nuclear and gas. The first has a more negligible cost and carbon footprint than the latter. Nuclear generation is relatively more constant in price and emissions since its production is more stable, though sometimes it is insufficient to supply all the houses. Hence they decide to produce energy with gas which is more expensive than nuclear energy and emits more carbon emissions (refer to \ref{fig:env_price_emissions}). 

\subsection*{Dataset generation}
The dataset generated for this problem has the following parameters (all of them normalized):
\begin{itemize}
    \item The demand profiles (family, teenager, business) explained before.
    \item Peak load maximum: The maximum the house can consume. 
    \item PV (photovoltaic) peak generation: The maximum is possible to generate with the solar panels for that house.  
    \end{itemize}
Battery characteristics
\begin{itemize}
    \item Random state of charge: to decide if the battery will start with a random percentage. 
    \item Capacity: energy capacity of the battery/array of batteries in (kWh) when is not normalized.
    \item Efficiency:  A value between 0 and 1 represents the one-way efficiency of the battery, considering the same efficiency for charging and discharging (\%).
    \item State of charge (SoC) max and min: Value between 0 and 1 representing the highest value and the lowest the SoC a battery can reach.
    \item P charge max: Maximum charging rate of a battery (\%).
    \item P discharge Max:  Minimum battery charging rate (\%).
    \item Sell Price: Price for injecting energy into the battery (reward to the prosumers).
    \item Buy price: Price for using energy from the battery (\$/kWh).
\end{itemize}   
The configurations for the training, evaluating, and testing in the project are found in Figure \ref{tab:conf_a2c_train}, \ref{tab:conf_a2c_eval} and \ref{tab:conf_a2c_test}. As demonstrated, for train and evaluation, the microgrids are of 6 houses, but for testing, there are 10. In the current version, the RL algorithm worked with multiple houses (microgrid) simultaneously and, before, worked with only one.

\begin{figure}[ht]
    \centering
        \includegraphics[width=\textwidth]{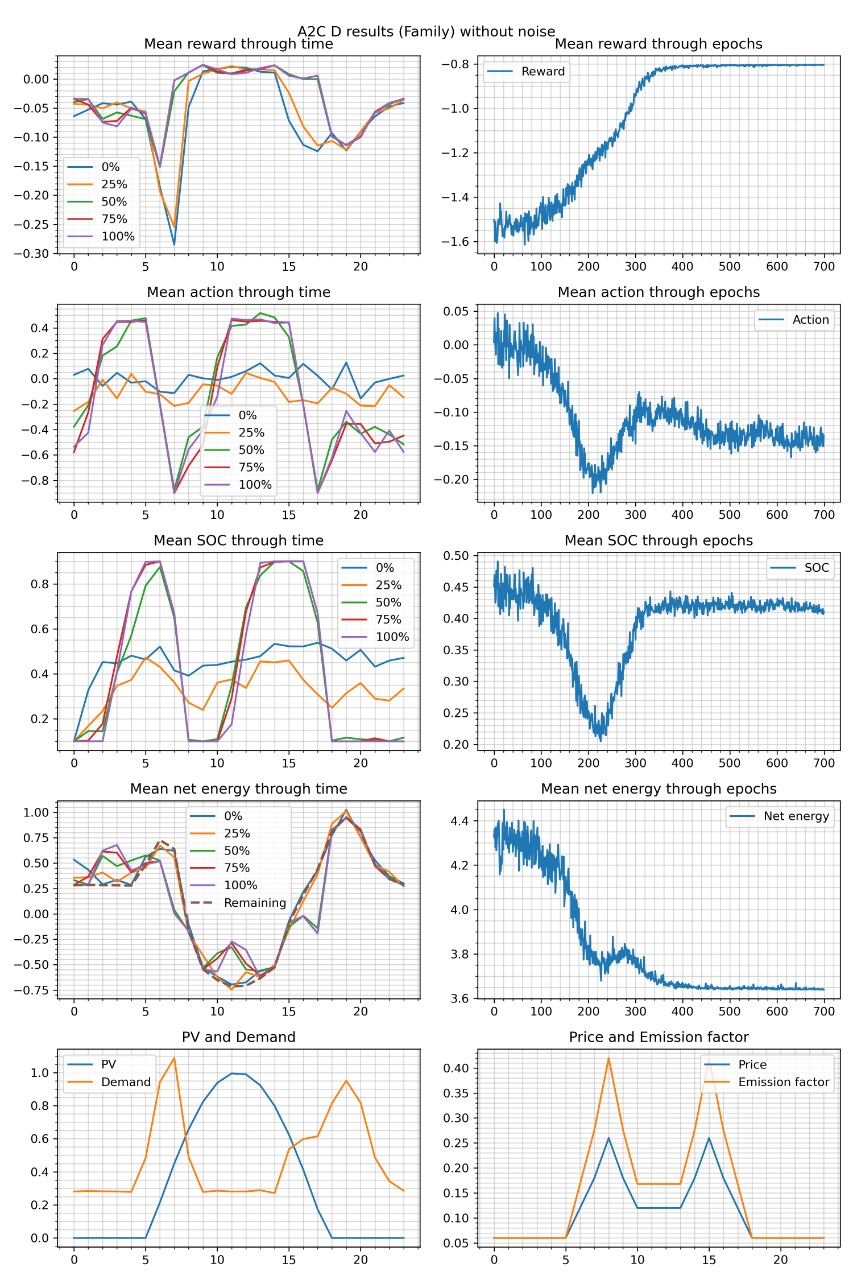}
        \caption{Results of the A2C with the dataset that has no noise.}
        \label{fig:fam_without_noise} 
\end{figure}

\begin{figure}[ht]
        \includegraphics[width=\textwidth]{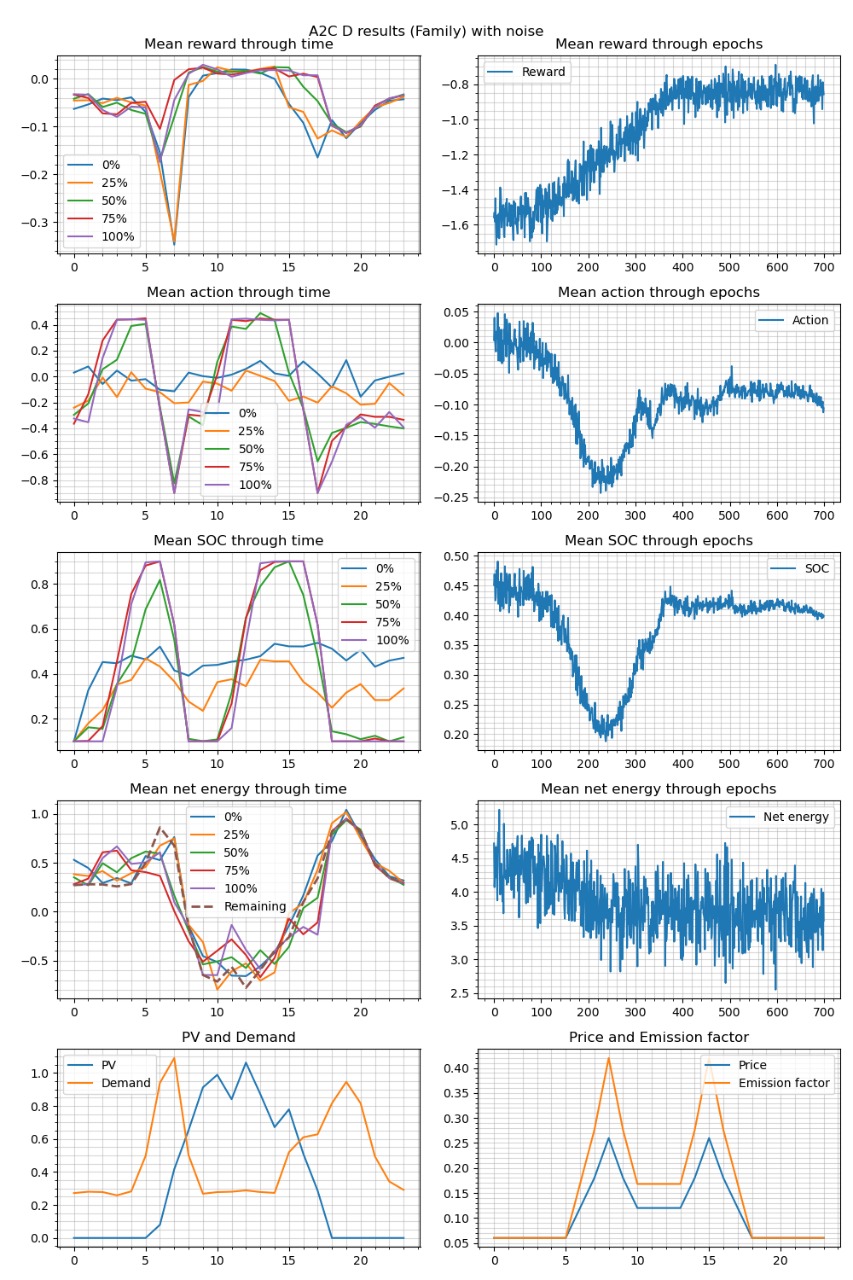}
    \caption{Results of the A2C with dataset that has noise.}
    \label{fig:fam_with_noise} 
\end{figure}

The data generated for the demand of the different houses are based on the main pattern for each profile, nonetheless what changes between homes are the state of the battery and the generation of energy with the solar panels (PV), which is not the same because of the incorporation of the noise (shown in Figure \ref{fig:fam_without_noise} and \ref{fig:fam_with_noise} in the subfigure ``PV and Demand'') for the generation and the energy load, modeled both of them using the Gaussian distributions $\mathcal{N}_{\text{pv}}(0,0.1)$ and $\mathcal{N}_{\text{load}}(0,0.01)$. Solar energy generation takes a sine function, shifting it to start after 5 am and shortening it to mimic the morning/daylight. After that, we incorporate the noise to replicate the possible clouds or weather conditions that can be present. The noise shows that there is a different result in the mean net energy through time (shown in Figure \ref{fig:fam_without_noise} and \ref{fig:fam_with_noise} in the subfigure ``Mean net energy through time'').

As shown in the tables above, some houses have no solar energy production (the ones in 0's), which means they need to rely on the battery to make decisions related to the energy. There is also no battery cell price so far, but this is one of the parameters planned to be incorporated in the following steps to see more dynamics in the microgrid.

\subsection*{Hyperparameters}

In the table \ref{tab:pg_hyperparameters}, we defined the following hyperparameters for the training after fine-tuning using grid search. Since there is less variance in the Advantage Actor-Critic (A2C), the number of epochs needed is less than using a policy gradient (PG). 

\begin{table}[tb]
    \centering
    \begin{tabular}{l c c}
        \\ \hline
        & PG     & A2C    \\ \hline
        \multicolumn{1}{l}{Number of discrete actions}  & 40     & 40     \\ \hdashline
        \multicolumn{1}{l}{Learning rate of the actor}  & 0.00381 & 0.00245 \\ \hdashline
        \multicolumn{1}{l}{Hidden layers of the actor}  & 128    & 128    \\ \hdashline
        \multicolumn{1}{l}{Learning rate of the critic} & -      & 0.001 \\ \hdashline
        \multicolumn{1}{l}{Hidden layers of the critic} & -      & 128    \\ \hdashline
        \multicolumn{1}{l}{Discount factor}             & 1.0    & 1.0    \\ \hdashline
        \multicolumn{1}{l}{Batch size}                  & 32     & 32     \\ \hdashline
        \multicolumn{1}{l}{Roll-out steps}              & 24     & 24     \\ \hdashline
        \multicolumn{1}{l}{Training steps}              & 2000   & 2000    \\ \hline
    \end{tabular}
    \caption{Hyper-parameter configuration for RL algorithms of the first layer.}
    \label{tab:pg_hyperparameters}
\end{table}

\begin{table}[tb]
    \begin{tabular}{l c c c c c c}
        \\ \hline
        & $house_1$ & $house_2$ & $house_3$ & $house_4$ & $house_5$ & $house_6$ \\ \hline
        \multicolumn{1}{l}{profile\_type}  & family & business & teenagers & family & business & teenagers     \\ \hdashline
        \multicolumn{1}{l}{profile\_peak\_load}  & 1 & 1 & 1 & 0.5 & 0.3 & 0.2 \\ \hdashline
        \multicolumn{1}{l}{battery\_random\_soc\_0} & False & False & False & False & False & False \\ \hdashline
        \multicolumn{1}{l}{battery\_capacity} & 1 & 1 & 1 & 1 & 1 & 1 \\ \hdashline
        \multicolumn{1}{l}{battery\_efficiency} & 1 & 1 & 1 & 1 & 1 & 1 \\ \hdashline
        \multicolumn{1}{l}{battery\_soc\_max} & 0.9 & 0.9 & 0.9 & 0.9 & 0.9 & 0.9 \\ \hdashline
        \multicolumn{1}{l}{battery\_soc\_min} & 0.1 & 0.1 & 0.1 & 0.1 & 0.1 & 0.1 \\ \hdashline
        \multicolumn{1}{l}{battery\_p\_charge\_max} & 0.8 & 0.8 & 0.8 & 0.8 & 0.8 & 0.8 \\ \hdashline
        \multicolumn{1}{l}{battery\_p\_discharge\_max} & 0.8 & 0.8 & 0.8 & 0.8 & 0.8 & 0.8\\ \hdashline
        \multicolumn{1}{l}{pv\_peak\_pv\_gen}  & 1 & 1 & 1 & 0.0 & 1 & 0.6 \\ \hline
    \end{tabular}
    \caption{Configuration of the houses in the microgrid for Training A2C.}
    \label{tab:conf_a2c_train}
\end{table}

% ----------------------------- Evaluating A2C
% \begin{figure}[ht]
%     \includegraphics[width=\textwidth]{img/c_a2c_mg_eval_houses.png}
%     \caption{Configuration of the houses in the microgrid for Evaluating A2C.}
%     \label{fig:conf_a2c_eval} 
% \end{figure}
\begin{table}[tb]
    \begin{tabular}{l c c c c c c}
        \\ \hline
        & $house_1$ & $house_2$ & $house_3$ & $house_4$ & $house_5$ & $house_6$ \\ \hline
        \multicolumn{1}{l}{profile\_type}  & family & business & teenagers & family & business & teenagers     \\ \hdashline
        \multicolumn{1}{l}{profile\_peak\_load}  & 1 & 0.8 & 0.5 & 0.2 & 0.3 & 0.2\\ \hdashline
        \multicolumn{1}{l}{battery\_random\_soc\_0} & False & False & False & False & False & False \\ \hdashline
        \multicolumn{1}{l}{battery\_capacity} & 1 & 1 & 1 & 0.5 & 0.9 & 0.9 \\ \hdashline
        \multicolumn{1}{l}{battery\_efficiency} & 1 & 1 & 1 & 1 & 1 & 1 \\ \hdashline
        \multicolumn{1}{l}{battery\_soc\_max} & 0.9 & 0.9 & 0.9 & 0.9 & 0.9 & 0.9 \\ \hdashline
        \multicolumn{1}{l}{battery\_soc\_min} & 0.1 & 0.1 & 0.1 & 0.1 & 0.1 & 0.1 \\ \hdashline
        \multicolumn{1}{l}{battery\_p\_charge\_max} & 0.8 & 0.8 & 0.8 & 0.8 & 0.8 & 0.8 \\ \hdashline
        \multicolumn{1}{l}{battery\_p\_discharge\_max} & 0.8 & 0.8 & 0.8 & 0.8 & 0.8 & 0.8\\ \hdashline
        \multicolumn{1}{l}{pv\_peak\_pv\_gen}  & 0.5 & 1 & 0 & 1 & 0.3 & 0.6 \\ \hline
    \end{tabular}
    \caption{Configuration of the houses in the microgrid for Evaluating A2C.}
    \label{tab:conf_a2c_eval}
\end{table}

% ----------------------------- Testing A2C
% \begin{figure}[ht]
%     \includegraphics[width=\textwidth]{img/c_a2c_mg_test_houses.png}
%     \caption{Configuration of the houses in the microgrid for Testing A2C.}
%     \label{fig:conf_a2c_test} 
% \end{figure}
\begin{table}[tb]
    \centering
    % width=.5\textwidth,center,caption=mytable,float=table
    \begin{adjustbox}{width=1.37\textwidth, center}
    \begin{tabular}{l c c c c c c c c c c}
        \\ \hline
        & $house_1$ & $house_2$ & $house_3$ & $house_4$ & $house_5$ & $house_6$& $house_7$ & $house_8$ & $house_9$ & $house_{10}$ \\ \hline
        \multicolumn{1}{l}{profile\_type}  & family & business & teenagers & family & business & teenagers  & family & business & teenagers & family    \\ \hdashline
        \multicolumn{1}{l}{profile\_peak\_load}  & 1 & 1 & 1 & 0.2 & 0.6 & 0.4 & 0.4 & 1 & 0.1 & 1\\ \hdashline
        \multicolumn{1}{l}{battery\_random\_soc\_0} & False & False & False & False & False & False & False & False & False & False \\ \hdashline
        \multicolumn{1}{l}{battery\_capacity} & 1 & 1 & 1 & 1 & 1 & 1 & 0.8 & 0.2 & 1 & 0.2 \\ \hdashline
        \multicolumn{1}{l}{battery\_efficiency} & 1 & 1 & 1 & 1 & 1 & 1 & 1 & 1 & 1 & 1 \\ \hdashline
        \multicolumn{1}{l}{battery\_soc\_max} & 0.9 & 0.9 & 0.9 & 0.9 & 0.9 & 0.9 & 0.9 & 0.9 & 0.9 & 0.9  \\ \hdashline
        \multicolumn{1}{l}{battery\_soc\_min} & 0.1 & 0.1 & 0.1 & 0.1 & 0.1 & 0.1 & 0.1 & 0.1 & 0.1 & 0.1\\ \hdashline
        \multicolumn{1}{l}{battery\_p\_charge\_max} & 0.8 & 0.8 & 0.8 & 0.8 & 0.8 & 0.8 & 0.8 & 0.8 & 0.8 & 0.8\\ \hdashline
        \multicolumn{1}{l}{battery\_p\_discharge\_max} & 0.8 & 0.8 & 0.8 & 0.8 & 0.8 & 0.8 & 0.8 & 0.8 & 0.8 & 0.8\\ \hdashline
        \multicolumn{1}{l}{pv\_peak\_pv\_gen}  & 0 & 0 & 0 & 0.7 & 1 & 0.7 & 1 & 1 & 1 & 0 \\ \hline
    \end{tabular}
    \end{adjustbox}
    \caption{Configuration of the houses in the microgrid for Testing A2C.}
    \label{tab:conf_a2c_test}
\end{table}

\begin{figure}[ht]
    \includegraphics[width=\textwidth]{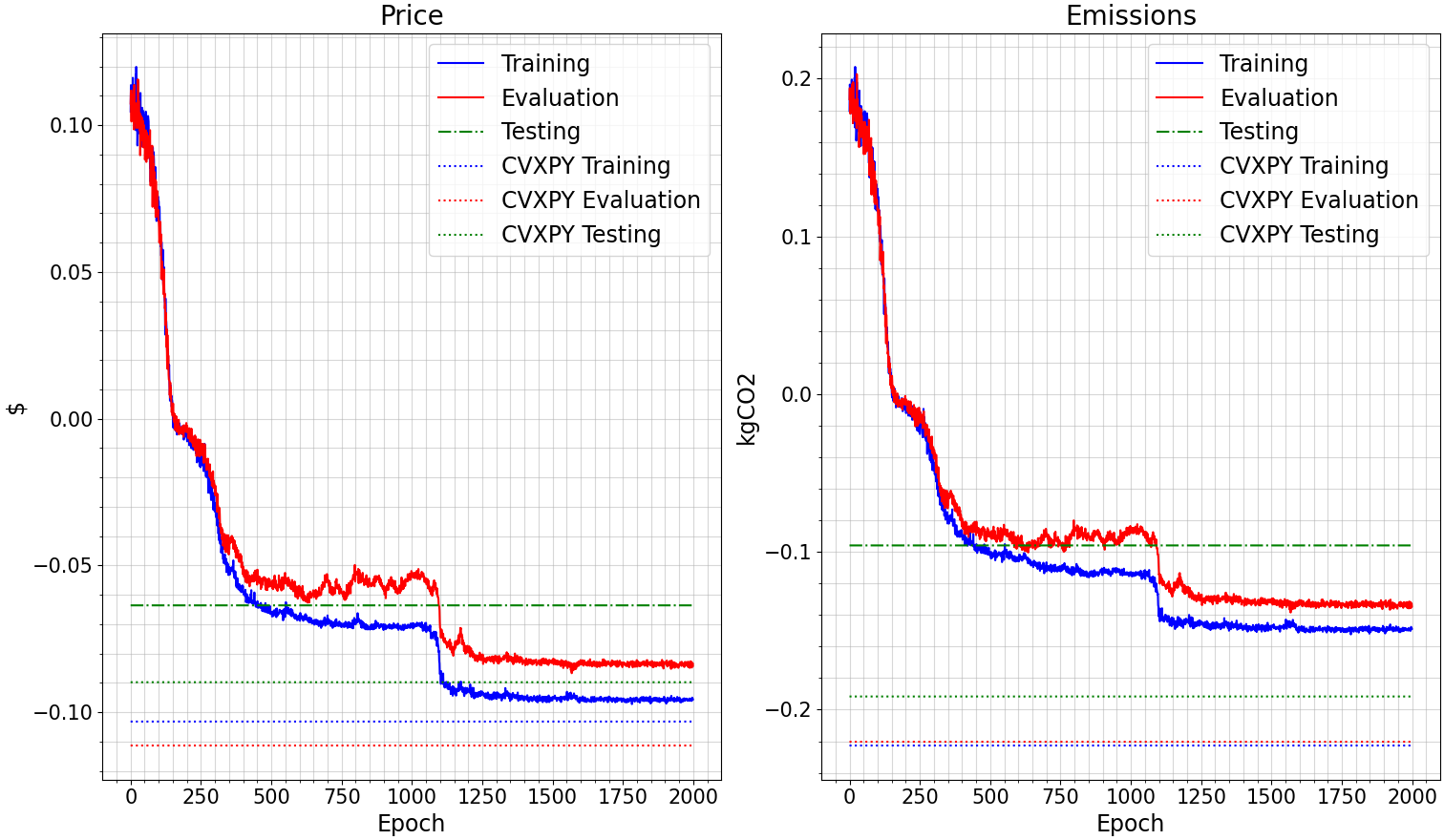}
    \caption{Emissions and Price score comparison between CVXPY solver and A2C in the microgrid for all 3 stages(training, evaluation, and testing).}
    \label{fig:cvx_optimize_metrics} 
\end{figure}

\end{document}